\documentclass{article}

\usepackage{cite}
\usepackage{amsmath,amssymb,amsfonts}
\usepackage{algorithm}
\usepackage{algorithmic}
\usepackage{graphicx}
\usepackage{textcomp}
\usepackage{authblk}
\usepackage{xcolor}
\begin{document}

\title{Video-based estimation of pain indicators in dogs}

\author[1]{Hongyi Zhu}
\author[2]{Yasemin Salgırlı}
\author[3]{Pınar Can}
\author[2]{Durmuş Atılgan}
\author[1,4]{Albert Ali Salah}
\affil[1]{Department of Information and Computing Sciences, Utrecht University, the Netherlands}
\affil[2]{Faculty of Veterinary Medicine, Department of Physiology, Ankara University, Turkey}
\affil[3]{Faculty of Veterinary
Medicine, Department of Surgery, Ankara University, Turkey}
\affil[4]{Department of Computer Engineering, Boğaziçi University, Turkey}
\maketitle

\begin{abstract}
Dog owners are typically capable of recognizing behavioral cues that reveal subjective states of their dogs, such as pain. But automatic recognition of the pain state is very challenging. This paper proposes a novel video-based, two-stream deep neural network approach for this problem. We extract and preprocess body keypoints, and compute features from both keypoints and the RGB representation over the video. We propose an approach to deal with self-occlusions and missing keypoints. We also present a unique video-based dog behavior dataset, collected by veterinary professionals, and annotated for presence of pain, and report good classification results with the proposed approach. This study is one of the first works on machine learning based estimation of dog pain state.
\end{abstract}

\section{Introduction}
%describe the problem 1-2 paragraphs
Interpretation of the behavior of animals is critical to understand their well-being. In this paper, we propose an approach for automatic estimation of pain states of dogs, using video modality. This problem is important for several potential applications, including long-term automatic monitoring of dogs, assessment and diagnosis aids for veterinary clinicians, and early warning systems for nonexperts. It is challenging, because dogs come in different breeds that have very different appearances, and exhibit different behavioral coping strategies. 

Animal pain estimation from video uses facial expressions of animals, as well as cues extracted from the body and the posture of the animal. While a comprehensive work for automatic dog pain estimation is missing from the literature, there are works on sheep, mice, pigs, and horses. Each animal requires a separate data collection effort, a different analysis approach, and different pain annotation schemes.

In this paper, we propose a pipeline for pain estimation in dogs, incorporating a number of off-the-shelf and in-house developed tools. We also introduce a database of dog videos, annotated by veterinary experts, to evaluate our approach. Pain estimation in dogs is an under-researched problem, and there are limited resources; we hope that our approach will establish a good baseline, and there will be more research to follow. We make all our code and annotations available publicly\footnote{The dataset, while not publicly available, can be accessed under special conditions permitted by the informed consent forms, such as federated learning approaches. Please contact the second author, if you are interested in the data.}.
%optionally - what does the literature say about it? what has been done before
%describe what the paper has to offer (contribution)
%optionally, the paper's structure
\section{Related Work}
Automated pain estimation for animals focuses on indicators from the body and the face of the animal. We first discuss pose estimation in this section, and then move to pain estimation. 
\subsection{Pose estimation}
\begin{figure}[htb]
\centering
\includegraphics[width=0.8\textwidth]{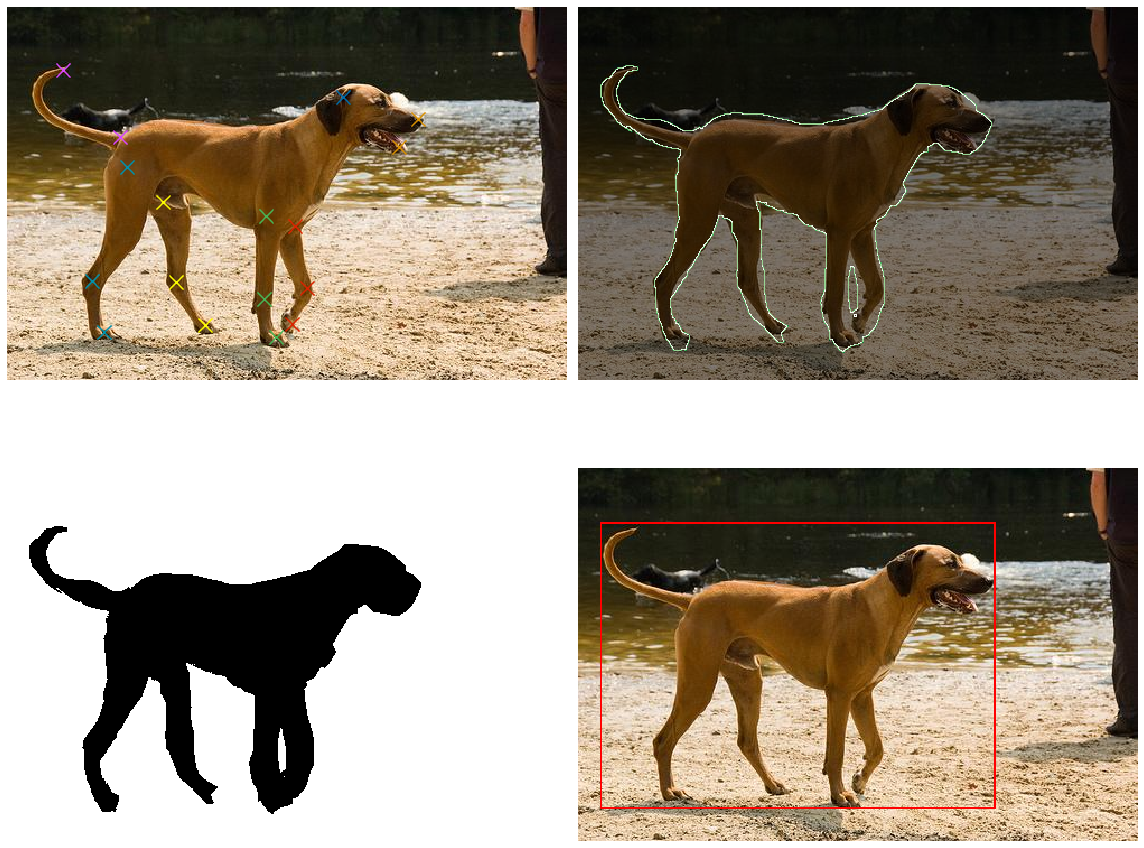}

\caption{\label{fig:representation}Image based representations for the dog. (1) Body key-joints, (2) Outline boundary, (3) Instance segmentation, (4) Bounding box}

\end{figure}

Pain estimation requires that the animal is properly detected and suitably represented. There are four main approaches for representing the dog, as shown in Figure~\ref{fig:representation}.

\begin{itemize}
\item Body key-joints representation: This representation allows easy modeling of pose semantics, and resembles the skeleton models popular in human behavior analysis. Examples of body joints are ``shoulders", ``middle joint of knees", ``paw", and ``spine". 

\item Outline boundary: In this representation, the boundary of the dog is marked. This method can represent the rigid shape of the animal.

\item Instance segmentation: Processing the image pixel-wise and segmenting the dog out of the background is called instance segmentation. This representation is used in visual scene understanding applications, and subsequently, there are several image databases that have this kind of ground truth annotations for dog images.

\item Bounding box: This representation is a cheaper-to-annotate version of instance segmentation that uses a rectangle to mark out the dog's extent in the image. Some pose estimation and instance segmentation algorithms would require bounding box detection as a first step to reduce background interference.
\end{itemize}

Body key-joints is the main representation method for pose estimation approaches, as they depict the geometrical configuration of multiple body parts. For a pose estimation algorithm, some methods would first use the bounding box to identify the location of a single animal object~\cite{papandreou2017towards, fang2017rmpe, chen2018cascaded} or apply instance segmentation to extract the shape of the animal body. Combining the advantage of the above four representation methods can help an algorithm keep track of the target animal and exclude noise from the background.

%animal pose estimation
Recent approaches for animal pose estimation like DeepLabCut~\cite{mathis2018deeplabcut} and LEAP~\cite{pereira2019fast} use a convolutional neural network (CNN) to directly extract features from animal images and to convert it to the key-joint representation. To train the CNNs, a human-annotated training set is used. To deal with the lack of large amounts of human-annotated data, transfer learning is the preferred solution. In DeepLabcut, a ResNet model~\cite{7780459} pre-trained on an object recognition task is used to enable learning animal specific estimators from only hundreds of annotated training samples. Additionally, in~\cite{Graving2019}, active learning is used to reduce the number of required annotations. In this approach, the system will suggest which images to annotate in each round, selecting these images actively. On the other hand, semi-supervised learning is used in~\cite{wu2020deep}, where a larger set of images without annotations is used jointly with a smaller set of annotated images to perform pose estimation. Apart from these three main approaches that can be combined (i.e. transfer learning, active learning, and semi-supervised learning), data augmentation is also used to increase the amount of available samples.

%Describe HRNet 
In this work, we use the Deep High-Resolution Network (HRNet), which is a state-of-the-art framework for 2D pose estimation that illustrated superior performance on a number of case studies~\cite{sun2019deep}. This approach uses an innovative network architecture, described in Section~\ref{sect:pose}. 

%Since our dog pain dataset has different background and dog movement. We measure the performance of pre-trained HRNet on our dog pain dataset and public dog dataset StanfordExtra \cite{biggs2020left}.

%information from human pose estimation
The pose estimation approaches we mentioned are generic and can be used with different animals. The DeepPoseKit method, for example, is applied on vinegar fly, locust, and zebra pose estimation problems~\cite{Graving2019}. However, popular human pose estimators use further information from the human skeleton structure~\cite{cao2017realtime}. In a similar way, animal-specific skeleton models can be used to improve the detection of the pose.

%dog-specific pose estimation
A number of approaches have been used or proposed for pose estimation in dogs and other canines. 
Modern object detection algorithms, trained on large sets, such as MS COCO~\cite{lin2014microsoft}, typically contain a ``dog" class. These can be leveraged to find a bounding box (or a more detailed segmentation) for detecting dogs in an image. 
 The work by Tsai and Huang \cite{9281102} introduced a multiple stages pose estimation algorithm, which uses a Mask R-CNN model \cite{8237584} to generate the contour mask map of the dog. A faster R-CNN \cite{girshick2015fast} performs posture recognition on the contour mask image and key part recognition on the original image. The key points of the skeleton would be obtained by analysing the candidate bounding box and animal pose jointly. 

In the RGBD-Dog approach \cite{Kearney_2020_CVPR} an RGBD dog dataset with landmark ground truth is used, generated from a 3D motion capture system. They applied a stacked hourglass network \cite{newell2016stacked} to predict a set of 2D heatmaps for a given depth image. This can determine the 3D coordinates of the body joints. A Hierarchical Gaussian Process Latent Variable Model (H-GPLVM) \cite{lawrence2003gaussian} is applied to refine the predicted 3D joint positions and to prevent pose ambiguities. However, this method requires depth data, which is not always available.

In this work, we use the HRNet-W32 model obtained from \cite{mmpose2020}, where W32 denotes the width of the high resolution network in the last three layers. The model is initially trained on the AnimalPose dataset~\cite{Cao_2019_ICCV}, which contains pose annotation on five animal categories (i.e. dog, cat, cow, horse, sheep, respectively) with 6,000+ instances in 4,000+ images, and 20 body keypoints annotated. We will describe the experiments in Section~\ref{sec:experiments}.
\subsection{Pain estimation}
%Pain is complex, we can just look at its indicators, and in dogs, these are the indicators:
Pain is defined as “an unpleasant sensory and emotional experience associated with, or resembling that associated with, actual or potential tissue damage,”~\cite{raja2020revised}. It is not possible to measure the subjective experience of pain in animals from a simple image, or a video, but it is possible to automatically observe indications that there is pain, and quantify them to some extent to create useful applications, such as early warning or long-term monitoring systems. It is noted that behavioral cues are not very reliable for pain estimation, because the animal can choose not to exhibit behaviors under certain circumstances.

Automatic, computer vision based pain estimation has not been extensively researched for dogs, but there is important work for other animals, such as horses~\cite{hummel2020automatic,pessanha2022facial},  mice~\cite{andresen2020towards}, rabbits~\cite{Keating2012EvaluationResponses} and sheep~\cite{Pessanha2020TowardsVideo} (see~\cite{broome2022going} for a recent survey). These works mostly focus on the face of the animal, and use validated clinical scales that rely on a series of observations to score the presence of pain indicators. These observations can include the position of the ears, the visibility of sclera in the eyes, muscle tension in certain areas like the mouth, or presence of specific behaviors, such as baring the teeth in horses, which are exhibited when the animal is pain.

There is some related work on emotion estimation for dogs, which is relevant as a basis for pain estimation. In~\cite{franzoni2019preliminary}, three emotional states (i.e. growl, sleep, and smile, respectively) are automatically estimated from sequences of dog images, based on ears, eyes, mouth and head features. In~\cite{ferres2022predicting}, DeepLabCut is used to estimate anger, fear, happiness, and relaxation in dogs, based on a custom dataset of 400 images, evenly distributed over these four classes.

The main challenges for computer vision based estimation of dog pain are the existence of many different dog breeds, which implies different sizes and colors that change the appearance, the presence of subtle cues, such as muscle contractions that are hidden beneath the coat, and the difficulties that plague ordinary face and body analysis, such as pose and illumination variations. Most importantly, there is a lack of data with pain annotations.

\section{Methodology}
\begin{figure}[thb]
\centering
\includegraphics[width=0.8\textwidth]{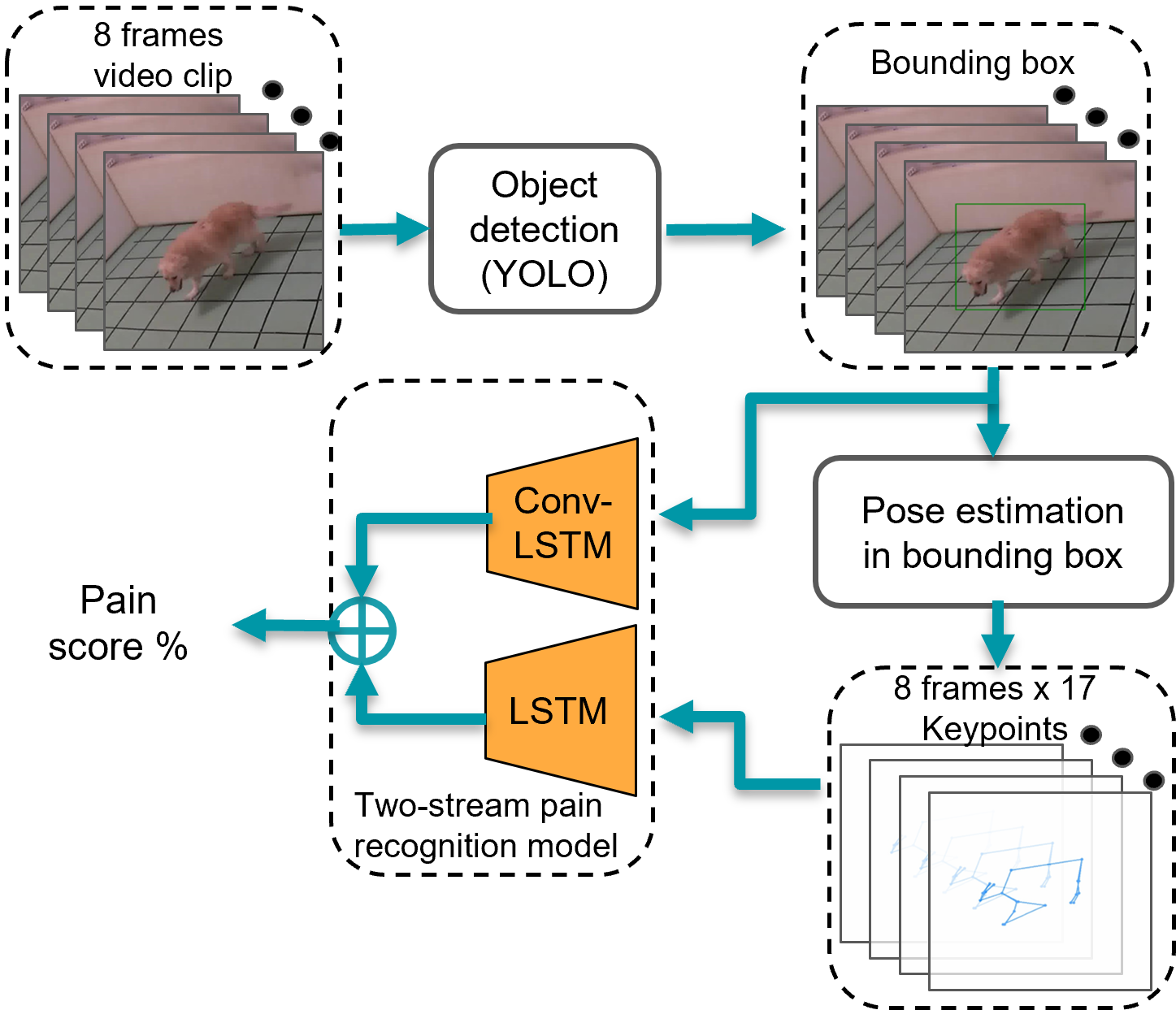}
\caption{\label{fig:pipline}Overall pipeline of the proposed method.}
\end{figure}

Our proposed approach for detecting the visual indication of a dog's pain state is made up of three parts. The off-the-shelf model for pose estimation is discussed in Section~\ref{sect:pose}. The feature extraction and  pre-processing are described in Section~\ref{sect:feature}, followed by pose normalization. In Section~\ref{sect:temporal}, we propose an architecture based on two-stream deep learning for pain detection. Figure~\ref{fig:pipline} depicts our proposed system, which takes a short, 8-frame video clip (sampled over 4 seconds) as input, estimates the dog's pose, and produces a pain score in the 0-1 range, which can be treated as a probability that the video contains a dog in pain.

%Be sure that the  symbols in your equation have been defined before or immediately following the equation. Use ``\eqref{eq}'', not ``Eq.~\eqref{eq}'' or ``equation \eqref{eq}'', except at  the beginning of a sentence: ``Equation \eqref{eq} is . . .''

%%  ---- START: Added 2022-03-15 for ACII2022 onwards -----

\subsection{Dog pose estimation}
\label{sect:pose}
%We use the HRNet, we will compare it with other baselines
For the initial detection of the dog in the images, we use a YOLOv5 deep neural network model \cite{redmon2018yolov3, redmon2017yolo9000}, pre-trained on the MSCOCO dataset \cite{lin2014microsoft}, which includes a ``dog" class. The RGB video frames are provided to the network as input, and the network outputs predicted bounding boxes, a confidence score, and the likelihood of the dog class. We set the confidence threshold to 0.25, and we select only the bounding boxes with a confidence score for containing a dog greater than the threshold.

\begin{figure}[htb!]
\centering
\includegraphics[width=0.8\textwidth]{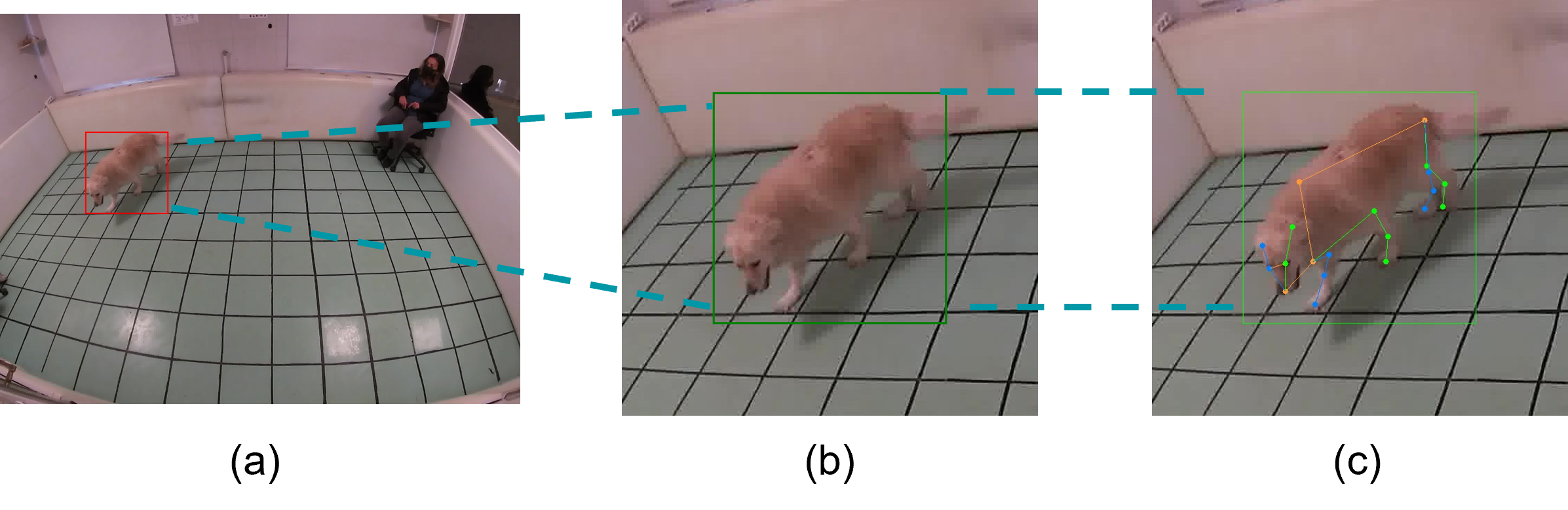}
\caption{\label{fig:bbox}Example of bounding box localization and keypoints extraction. (a) We first apply the YOLOv5 algorithm to detect the bounding box of the dog. (b) We then crop the image with the center of the bounding box to exclude noisy background information. The bounding box is extended by 10\% to ensure that the majority of the dog's body is included. (c) We apply the HRNet pose estimation algorithm to detect the body keypoints within the bounding box, and generate the coordinates of the keypoints.}
\end{figure}

By combining the feature of YOLOv5 and HRNet, we can construct a data pre-processing pipeline to extract both RGB image and posture information as the model input.  (See Figure~\ref{fig:bbox} for the example image pre-processing representation)

\begin{figure}[htb!]
\centering
\includegraphics[width=0.8\textwidth]{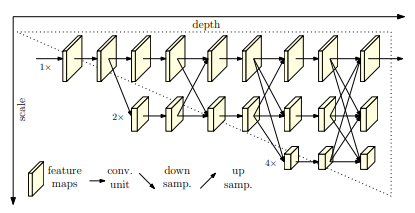}
\caption{\label{fig:HRNet}The structure of HRNet. The horizontal and vertical axes correspond to the depth of the network and the scale of the feature maps, respectively (Adapted from \cite{sun2019deep}).}
\end{figure}

As stated before, we use the HRNet \cite{sun2019deep} approach for pose estimation. The majority of existing pose estimation methods, such as SimpleBaseline \cite{xiao2018simple} and DeepLabCut \cite{Nath2019}, adapt an encoder-decoder structure composed of a series of convolutional networks to downsample the input image from high to low-resolution feature maps and then apply transposed convolution layers to recover the original high-resolution feature map. HRNet, on the other hand, is composed of a central stem network that maintains high resolution throughout the process, as well as sample branch networks in each layer of the stem, and parallel connections between the multi-resolution branch networks. Additionally, HRNet implements multi-scale fusion between each parallel branch network, in order to facilitate information exchange throughout the process. The stem network's last layer would output keypoint estimates (see Figure~\ref{fig:HRNet}).

Since HRNet is a top-down pose estimation algorithm, it requires the approximate location of the target object in the image and preferably uses closely cropped images as inputs. The bounding box output of the YOLOv5 model is extended by 10\% to ensure we include the entire dog, and provided to HRNet for further processing. We use the HRNet-W32 model obtained from \cite{mmpose2020}, trained on the AnimalPose dataset~\cite{Cao_2019_ICCV}, at this stage. Figure~\ref{fig:bbox} illustrates the processing of a frame for bounding box localization and keypoints extraction.

\subsection{Feature extraction and pre-processing}
\label{sect:feature}
We extract 17 keypoints from various parts of the body of the dog, from the head, spine, and leg joints. The head includes two eyes, and the nose tip (3 points), spine includes withers and tail end (2 points) and legs include four elbows, four knees, and four paws (12 points).
During training, since self-occlusions of joints is a common problem, we adopt a 
missing data treatment approach to augment some missing landmarks and to exclude some poorly  estimated images. The poses are centered and scaled to be consistent with one another. This ensures that our approach is not affected by the differing size of the dog on the image (due to different distances to the camera).

Problems due to occlusions and self-occlusions can be partially solved by using body symmetry and by inference based on observed keypoints over time. The central concepts are as follows:
\begin{enumerate}
            \item If some keypoints are missing from a frame, they may be discovered and interpolated by using the neighbouring frames in the sequence. Note that this makes would only work in an ``offline" mode.

            \item If -after the first step- there are more than nine keypoints missing, or the spine keypoints are missing, the frame is discarded from further processing.
            
            \item Each leg has three keypoints (i.e. elbow, knee, and paw). If one keypoint of a leg is missing, it can be inferred according to the position of the other two keypoints and the spine points. 
            
            \item If more than one keypoint is missing in one leg, they can be guessed by exploiting the left and right side body symmetry, or front and back side symmetry.
\end{enumerate}
This approach for resolving missing data is simple and effective, and necessary for subsequent processing of the body pose. In comparison to the head and spine keypoints, the leg keypoints convey more detailed information about the pose. Thus, the essence of this approach is to compensate for the missing keypoints induced by self-occlusion. 

Our pose estimate step defines a mapping from the RGB image to body keypoints graph as:
\begin{equation}
s_{i}(t) \stackrel{\text { pose detector }}{\longrightarrow} p_(t) \quad \forall t=1, \ldots, T_{i},
\end{equation}
where $p_{i}(t)$ denotes the posture for sample $i$ at time frame $t$ and $T$ denotes the total sequence length for sample $i$. The whole dataset contains $N$ samples. In particular, $p_{i}(t)$ consists of a sequence of 2D coordinates as:
\begin{equation}
p_{i}(t)=\left\{\left(x_{j}(t), y_{j}(t)\right)_{i}\right\}_{j \in J},
\end{equation}
where $j$ is the index of keypoints, and $J$ is the number of keypoints defined in the pose estimation model. %In this paper $J=\{1, \ldots, 14\}$.
%AAS: Why 14 keypoints? We have 17 in the previous sections?
\subsection{Pose normalization}
%Synchronization
The pose normalization step aims to transform the keypoint coordinates from an absolute to a local root point-centered reference system. The transformed pose coordinates are defined as:
\begin{equation}
\label{eq: coord}
\left(\bar{x}_{j}, \bar{y}_{j}\right)_{i}=\left(x_{j}, y_{j}\right)_{i}-\left(x_{root}, y_{root}\right)_{i} \quad \forall j \in J
\end{equation}
The dependence of $t$ is omitted here for simplicity. In this paper, we use the middle point between the tail end and the neck as the root point. Thus, the root point centered coordinate is defined as:
\begin{equation}
\bar{p}_{i}=\left\{\left(\bar{x}_{j}, \bar{y}_{j}\right)_{i}\right\}_{j \in J}
\end{equation}
Furthermore, in order to normalize the size of the dog within the image, we re-scale the keypoint coordinates $\bar{p}_{i}$:
\begin{equation}
\overline{\bar{p}_{i}}=\frac{\bar{p}_{i}}{\left\|v_{neck, tail end}\right\|},
\end{equation}
where $\bar{v}_{neck, tail end}$ is the vector between the neck and the tail end landmarks.

The transformation and normalization of landmarks has two purposes: The placement of the dog within the image has no effect on the processing, and the distance of the dog to the camera, as well as its size become less important. One may assume that in case much more training data become available in the future, some clustering in this space, followed by cluster-specific models, may improve the modeling.
\subsection{Pain indicator estimation}
\label{sect:temporal}
\begin{figure}[htb]
\centering
\includegraphics[height=0.8\textwidth]{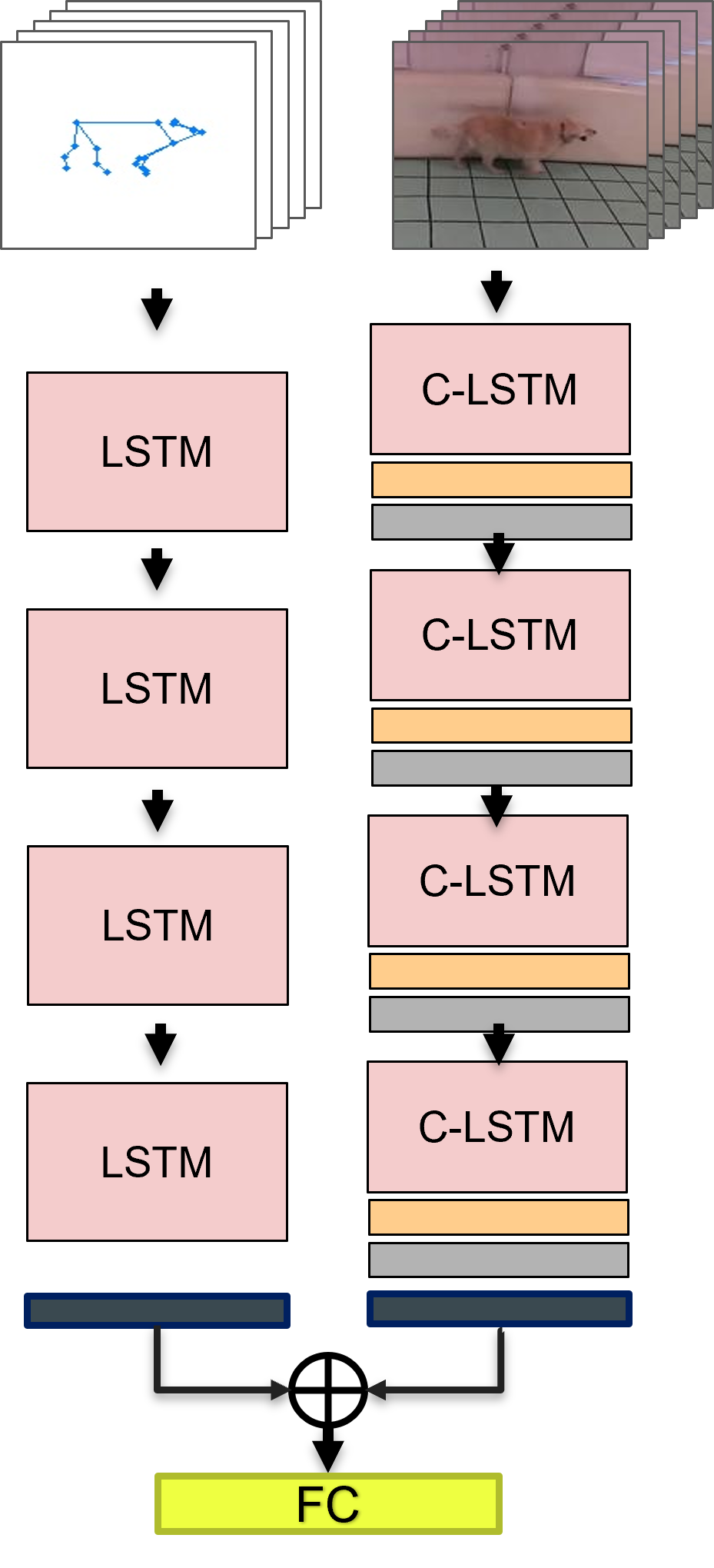}
\caption{\label{fig:net_fig} Our proposed  two-stream model for dog pain estimation. The left branch is a four layer LSTM model, which is responsible for processing the keypoint information. The right branch is a four layer convolutional LSTM model, which is responsible for processing RGB images. The yellow and gray blocks are the max pooling and batch normalization layers. Black blocks represent the time attention layers. The green block is the fully connected layer. We fuse our two branches by concatenating the last layers' feature vectors.}
\end{figure}

Based on several other works, such as \cite{karim2019multivariate, simonyan2014two, broome2019dynamics}, we propose to use a two-stream model structure to evaluate the temporal information in the detected pose representation.  In \cite{karim2019multivariate}, a two-stream model structure is used to predict human actions, one stream based on the long short term memory (LSTM) architecture~\cite{hochreiter1997long} and the other with a 1-dimensional convolutional network. Our proposed method resembles this structure (see Figure~\ref{fig:net_fig}), with some important differences. We use a basic LSTM branch to process the keypoint based pose representation, another, Convolutional LSTM based branch to process the RGB images. Both branches are complemented with an attention module before their outputs are combined with a fully convolutional layer, and the output is a binary classification of the pain state. We describe each component separately.

\subsubsection{Basic LSTM}
LSTM is a variant of recurrent neural networks, which use part of the internal information at time step $t$ in the processing of time step $t+1$. The basic LSTM network maintains a memory vector \text{m} to control the outputs at each time step. The equations for each time step are listed as follows:
\begin{equation}
\begin{aligned}
&\mathbf{i}_{t}=\sigma\left(\mathbf{W}^{u} \mathbf{h}_{t-1}+\mathbf{I}^{u} \mathbf{x}_{t}+\mathbf{b}_{i}\right) \\
&\mathbf{f}_{t}=\sigma\left(\mathbf{W}^{f} \mathbf{h}_{t-1}+\mathbf{I}^{f} \mathbf{x}_{t}+\mathbf{b}_{f}\right) \\
&\mathbf{o}_{t}=\sigma\left(\mathbf{W}^{o} \mathbf{h}_{t-1}+\mathbf{I}^{o} \mathbf{x}_{t}+\mathbf{b}_{o}\right) \\
&\mathbf{\tilde{c}}_{t}=\tanh \left(\mathbf{W}^{c} \mathbf{h}_{t-1}+\mathbf{I}^{c} \mathbf{x}_{t}+\mathbf{b}_{c}\right) \\
&\mathbf{c}_{t}=\mathbf{f}_{t} \odot \mathbf{c}_{t-1}+\mathbf{i}_{t} \odot \mathbf{\tilde{c}}_{t} \\
&\mathbf{h}_{t}=\tanh \left(\mathbf{o}_{t} \odot \mathbf{c}_{t}\right),
\end{aligned}
\end{equation}
where ${i}_{t}$, ${f}^{t}$, ${o}^{t}$, and ${c}^{t}$ are the activation gates of the inputs, forget vector, output, and cell states vector, respectively. $\tilde{c}_{t}$ is the cell input activation vector. ${h}_{t}$ is the hidden state of LSTM units, $\sigma$ represent the logistic sigmoid activation function, $\odot$ represents the elementary multiplication. The weight matrices for each gate are denoted as $\mathbf{w}^{u}$, $\mathbf{w}^{f}$, $\mathbf{w}^{o}$, $\mathbf{w}^{c}$. The projection matrices are defined as $\mathbf{I}^{u}$, $\mathbf{I}^{f}$, $\mathbf{I}^{o}$, $\mathbf{I}^{c}$. The bias vector parameters of the inputs, forget vector, outputs, and cell states vector are the ${b}_{i}, {b}_{f}, {b}_{o}, {b}_{c}$, which need to be learned during training.

The LSTM network is able to learn temporal information, and combine stored information from previous time steps into its prediction. It uses three gating functions to control this: a forgot gate, an input gate, and an out gate, whereby it deletes old information or adds new information into the current state. However, this algorithm is not able to distinguish which time step contains more crucial information, and it would be a challenge for LSTM to learn long-term dependencies from longer sequences. In order to address this problem, an attention mechanism was proposed in \cite{bahdanau2014neural} to learn long-term dependencies.
\subsubsection{Convolutional LSTM}
Traditional versions of LSTM would use the fully connected layers in every state transition and the gate output process. This kind of LSTM can also be called the \textsl{fully connected LSTM} (FC-LSTM). The FC-LSTM has the ability to extract temporal information from sequences, which makes it suitable for time series problems \cite{graves2013generating, hochreiter1997long, bahdanau2014neural, donahue2015long}. However, the major drawback of FC-LSTM is that its fully connected layers have too much redundancy for visual inputs. In order to address this problem, Shi et al. \cite{xingjian2015convolutional} introduced Convolutional LSTM (C-LSTM), which has convolutional structures in both the
input-to-state and state-to-state transitions. 

C-LSTM provides the ability to process the spatio-temporal sequence more effectively. A three-dimensional tensor can be fed into the network without flattening into a one dimensional tensor \cite{ranzato2014video, srivastava2015unsupervised}. All the inputs $x_{1},...,x_{t}$, cell outputs $c_{1},...,c_{t}$, hidden states $h_{1},...,h_{t}$, and gates $f_{t}, i_{t}, c_{t}$ of C-LSTM are 3D tensors, whose last two dimensions are spatial dimensions (rows and columns). This model can downsample the input image to extract high-dimensional features and capture the information across the video frames. We use multiple C-LSTM layers in our pipeline to process RGB frames in the video. The equations of C-LSTM are given in Eq.~\ref{eq:convlstm}, reproduced from \cite{xingjian2015convolutional}. 
\begin{equation}
\begin{aligned}
\label{eq:convlstm}
f_{t} &=\sigma_{g}\left(W_{f} * x_{t}+U_{f} * h_{t-1}+V_{f} \circ c_{t-1}+b_{f}\right) \\
i_{t} &=\sigma_{g}\left(W_{i} * x_{t}+U_{i} * h_{t-1}+V_{i} \circ c_{t-1}+b_{i}\right) \\
c_{t} &=f_{t} \circ c_{t-1}+i_{t} \circ \tanh\left(W_{c} * x_{t}+U_{c} * h_{t-1}+b_{c}\right) \\
o_{t} &=\sigma_{g}\left(W_{o} * x_{t}+U_{o} * h_{t-1}+V_{o} \circ c_{t}+b_{o}\right) \\
h_{t} &=o_{t} \circ \tanh\left(c_{t}\right)
\end{aligned}
\end{equation}
Here * denotes the convolution operator, and $\circ$ denotes the Hadamard product (element-wise product). $W, U, V$ and $b$ are the weight matrices and bias vector parameters that need to be learned during training.
\subsubsection{Attention LSTM}
In our model, we add attention operation on the top of the LSTM outputs. The attention mechanism adds a context vector $V$ onto the target sequence $y$. In \cite{bahdanau2014neural}, the elements of the context ${v}_{i}$ depend on a sequence of annotations (${b}_{1}$,..., ${b}_{T}$) with length of $T$. Each annotation would make a comparison across the whole input sequence and estimates which term of the input sequence should be paid more attention. The weighted sum of annotations ${b}_{i}$ is used to compute the context vector as follows:
\begin{equation}
v_{i}=\sum_{j=1}^{T_{x}} \alpha_{i j} b_{j}
\end{equation}
The score $\alpha_{i j}$ of each annotation is  inferred as:
\begin{equation}
\alpha_{i j}=\frac{\exp \left(e_{i j}\right)}{\sum_{k=1}^{T_{x}} \exp \left(e_{i k}\right)}
\end{equation}
The alignment model assigns a score $\alpha_{i j}$ to the pair of context vector elements at position $j$ and the annotation at position $i$. $e_{i j}$ is computed by the vector product between the LSTM hidden state ${v}_{i-1}$ and the $j^{th}$ annotation. In our model, both context vector and annotation are the hidden state sequences of the last LSTM layer, which forms a self-attention model. In \cite{bahdanau2014neural}, the alignment score $\alpha$ is parameterized by a feed-forward network with a single hidden layer, and this network is jointly trained with the other parts of the model.
\subsubsection{Fusion of branches}
The most direct way to combine the two feature vectors created by the two branches is concatenation fusion~\cite{feichtenhofer2016convolutional}. 
$\mathbf{y}^{\text{cat}} = f^{\text{cat}}\left(\mathbf{x}^{a}, \mathbf{x}^{b} \right)$ means we stack two feature vectors of the  feature channels $c$:
\begin{equation}
y_{c+c'}^{\text {cat }}=x_{c}^{a} \quad y_{c'}^{\text {cat }}=x_{ c'}^{b},
\end{equation}
where $\mathbf{y}^{\text{cat}} \in \mathbb{R}^{N \times (C+C')}$. This is denoted with FC in Figure~\ref{fig:net_fig}.

\section{Datasets}
\label{sect:data}
There are a few datasets with dog images available for training pose estimation models, but no datasets are available for assessing pain indicators. The dataset we introduce in this paper, is unique in that respect. We use two other datasets for assessing the pose estimation method that we use in this paper, namely, StanfordExtra \cite{biggs2020left} and TigDog \cite{delpero15cvpr}. 

StanfordExtra is an image-based dog dataset with annotated 2D keypoints, obtained by labeling part of the existing Stanford Dog Dataset \cite{khosla2011novel}, which consists of 20,580 dog images taken “in the wild” from 120 different dog breeds. The dataset contains various poses, with variation on environmental occlusions, interaction with humans or other animals, and partial views. For each image, there are 19 candidate body key points, but only the points that can be observed are labeled. We selected 6 dog breeds out of 120 randomly, and used 60 images for each breed for comparison experiments. The dog breeds are Dandie-Dinmont, Tibetan-terrier, Bluetick, Rhodesian-ridgeback, Brittany-spaniel, and Brabancon-griffon, respectively.
\begin{figure}[htb]
\centering
\includegraphics[width=0.9\textwidth]{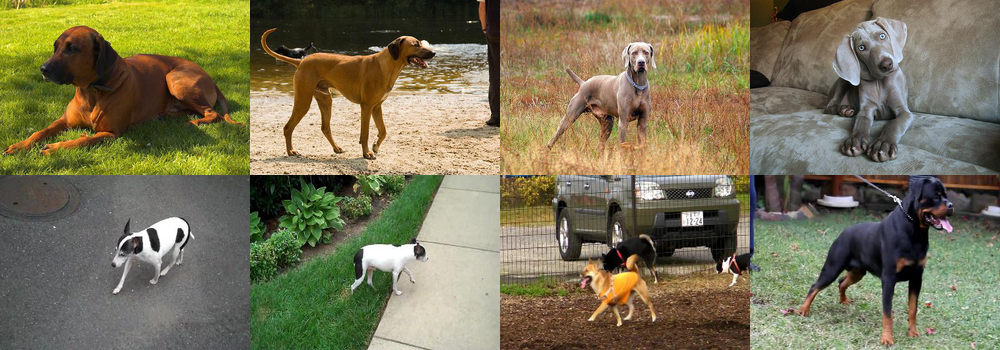}
%Hongyi, let's have less dogs per row, so that we can enlarge it a bit... How about just four dogs per dataset? I put here a poor quality version, please improve this...
\caption{\label{fig:examples}Example images from the StanfordExtra (top) and TigDog (bottom) datasets.}
\end{figure}

TigDog is a video base dataset with no annotated key-points, sourced from the Youtube-Objects dataset \cite{DBLP:conf/cvpr/PrestLCSF12}. For most videos, the camera is not static, and follows the dog, which is typically centered in the image. Figure~\ref{fig:examples} shows some example images from these two datasets. 

The Anonymous University Dog Behavior Dataset (AUDBD) used in our experiment is collected by the 
Veterinary Medicine Department of ANONYMOUS University, in a group that performs research into dog behavior
%~\cite{salgirlievaluation,salgirli20behavioral}. 
(citations removed for blind review).
The complete dataset includes 61 videos, each corresponding to an individual dog. The overall video duration is six hours and 45 minutes. There are 23 videos, totaling three hours and nine minutes, labeled as ``pain condition", and 38 videos with a total of three hours and 36 minutes labeled as ``non-pain" condition. 

Pain can be associated to wear and tear of tissues, injuries to the bones or a joint, or damage to muscles, ligaments, tendons or other soft tissues (musculoskeletal). These are marked as ``orthopedic pain" in the dataset. Other types of pain can be related to irritated or damaged nerves, and these are marked as ``neurological pain". The labels of the dataset are provided by the veterinarian experts according to their medical records. Every sample in the pain set is a real dog patient with a physical problem that could potentially cause pain symptoms, however, not all images in the videos will exhibit all pain indicators.

We extract and process the frames from the videos at two frames per second, and in clips of four seconds of video each. Each such clip is thus composed of eight frames. 
In order to have enough data to train our models, we use a two frame (i.e. one second) overlap between subsequent clips. 
This way, the dataset contains a total of 12,435 clips
%. The overall dataset scale is comparable to the other public action recognition datasets, such as UCF-101 \cite{soomro2012ucf101} and PoseTrack \cite{iqbal2017posetrack} 
(See dataset overview in Table~\ref{table:dataset}).

\begin{table}[htb!]
\centering
\begin{tabular}{|l|l|l|l|}
\hline
                & \# videos & \# clips & Duration \\ \hline
Orthopedic pain &   14        & 2,957         & 01:19:06         \\ \hline
Neurological pain & 9          & 2,161      & 01:50:08         \\ \hline
Total pain      & 23        & 5,118     & 03:09:14 \\ \hline
Not pain        & 38        & 7,584     & 03:36:50 \\ \hline
\end{tabular}
\caption{The overall statistics of the AUDBD  dataset. Frames are sampled at two FPS, and clips consist of eight frames with two frames overlap. Duration is shown in hh:mm:ss.}
\label{table:dataset}
\end{table}

%Noise conditions and limitations of the database

%Figure~\ref{fig:samples} shows examples from the three databases used in the paper.

\section{Experimental Results}
\label{sec:experiments}
\subsection{Pose estimation}
%We compare HRNet with DeepLabCut
We measure the performance of pre-trained HRNet on our dog pain dataset, as well as StanfordExtra and Tigdog. We use the Percentage of Correct Keypoints (PCK) measure, which is the percentage of keypoints detected falling within a normalized distance of the manually annotated ground truth. We use the square root of the bounding box area as the normalization index. The results are shown in Table~\ref{table:HRNet}.
\begin{table}[htb]
\centering
\begin{tabular}{lllll}
\hline
Dataset       & Head & Spine & Legs & Total \\ \hline
StanfordExtra & 93.4 & 88.9  & 84.0 & 86.5  \\ \hline
Tigdog        & 85.1 & 83.3  & 78.5 & 81.5  \\ \hline
AUDBD     & 88.4 & 82.1  & 77.3 & 80.9  \\ \hline
\end{tabular}
\caption{10 percent PCK (Percentage of Correct Keypoints) on three datasets. Head includes two eyes, and the nose tip. Spine includes withers and tail end. Legs includes four elbows, four knees, and four paws.}
\label{table:HRNet}
\end{table}

We can see that the pre-trained HRNet has a good performance in general, and has better performance on the StanfordExtra dataset compared to Tigdog and AUDBD, since StanfordExtra is visually more similar to the Animalpose dataset, on which the HRNet was trained.

%Tables 4.2 and 4.3 from the thesis.

\subsection{Pain estimation}
%Table 4.4.
Our classification models are evaluated using five fold cross-validation. This means that for a given video dataset, we create a video for each dog and then use 13$\%$ of the dogs as the test set and the remaining dogs for five-fold cross-validation. For each fold, we randomly sample $80\%$ of videos from the pain class and $80\%$ from the non-pain class to create a training set; the remaining videos would be used to create a validation set for early stopping. There is no dog used in multiple folds. The test set would not be used to train or validate the model. 

We contrast several models under this setting. We report the performance of LSTM and C-LSTM branches separately, where the former uses the keypoints, and the latter uses the RGB images. LSTM models are reported under three different settings: LSTM32, LSTM64, and LSTM128, respectively, where the number denoted the number of hidden units. For one-stream networks, we also report results with an I3D network \cite{carreira2017quo} as an additional baseline. Finally, we also report results with a frequently used dual-stream approach, where one branch processes RGB image, and the second branch processes the optical flow image of the same video. 
Table~\ref{table:pain} reports the average accuracy and F1 scores for these models obtained after five fold cross-validation, reported on the test set. The F1-score (i.e. the harmonic mean of precision and recall) is reported in addition to accuracy, since it is informative when the dataset is imbalanced. The model with the highest average F1 score and accuracy is C-LSTM+LSTM128, which has an average F1 score of $76.3\%\pm~4.4$ and an average accuracy of $77.0 \%\pm4.6$.
\begin{table}[tbh!]
\centering
%\adjustbox{max width=\textwidth}{%
\begin{tabular}{|l|l|ll|l|}
\hline
\textbf{One stream models} &  Data type                & \multicolumn{1}{l|}{Avg. F1} & Avg. Acc. &  \#Param. \\ \hline
C-LSTM                   & RGB              &  \multicolumn{1}{l|}{62.9$\pm$3.3}   & 64.5$\pm$3.6        & 8,105,986      \\  \cline{1-1}
LSTM64                  & KP              & \multicolumn{1}{l|}{58.8$\pm$5.5}   & 59.1$\pm$6.8         & 137,922     \\ \cline{1-1}
I3D                    & RGB              & \multicolumn{1}{l|}{74.1$\pm$2.5}   &74.6$\pm$1.1         & 24,327,632       \\ \hline
\textbf{Two stream models} &                  &  \multicolumn{2}{l|}{}                      &              \\ \hline
C-LSTM+LSTM32            & RGB+KP    &  \multicolumn{1}{l|}{70.3$\pm$3.3}   &  71.2$\pm$3.1        & 8,143,204      \\  \cline{1-1}
C-LSTM+LSTM64           & RGB+KP    &  \multicolumn{1}{l|}{72.9$\pm$3.5}   & 72.1$\pm$4.1        & 8,243,908   \\ \cline{1-1}
C-LSTM+LSTM128          & RGB+KP     &  \multicolumn{1}{l|}{76.3$\pm$4.4}   & 77.0$\pm$4.6          & 8,635,780    \\ \cline{1-1}
Double C-LSTM            & RGB+OF &  \multicolumn{1}{l|}{67.7$\pm$5.1}   & 69.2$\pm$3.9         &  16,210,403            \\ \hline
\end{tabular}
%}
\vspace{.1in}
\caption{Results ($\%$F1-score and accuracy) for different  models. RGB: images, KP: keypoints, OF: optical flow. 32, 64, 128 denote the number of hidden units in the LSTM models, which are used to process the keypoints sequence.}
\label{table:pain}
\end{table}
\subsection{Qualitative Analysis}
We use the Grad-CAM method~\cite{selvaraju2017grad} to compute activation (saliency) maps for our best model (i.e. ConvLSTM+LSTM) to visualize the model's attention. We apply the Grad-CAM method on the last convolutional layer of the RGB stream. In order to obtain the class discrimination saliency map, we first compute the gradient score of the class $c$, $y^{c}$, with respect to the feature map activations $A^{k}$ (layer output) in the last convolutional layer, i.e.\ $\frac{\partial y^{c}}{\partial A^{k}}$. The gradient flows are then processed by the global average pooling in the width and height dimensions (denote as $i$ and $j$ respectively) to get the neuron importance weights $\alpha_{k}^{c}$.
\begin{equation}
\alpha_{k}^{c}=\overbrace{\frac{1}{Z} \sum_{i} \sum_{j}}^{\text {global average pooling }} \underbrace{\frac{\partial y^{c}}{\partial A_{i j}^{k}}}_{\text {gradients via backprop}}
\end{equation}
where $Z$ is the product of width and height. The importance of feature maps (layer output channels) $k$ for a class $c$ are then weighted according to their corresponding gradient magnitude. The feature maps are processed by a RELU function to focus on the features which have a positive impact on the class of interest. 
\begin{equation}
L_{\mathrm{Grad}-\mathrm{CAM}}^{c}=\operatorname{ReLU} \underbrace{\left(\sum_{k} \alpha_{k}^{c} A^{k}\right)}_{\text {linear combination}}
\end{equation}
The region with the highest magnitude is considered to be the most important for the classification decision. The output feature map is up-sampled with an interpolation to the size of the original image, and superposed on it as an heatmap. 

\begin{figure}[htb!]
\centering
 \begin{tabular}{l}
  \includegraphics[width=0.8\textwidth]{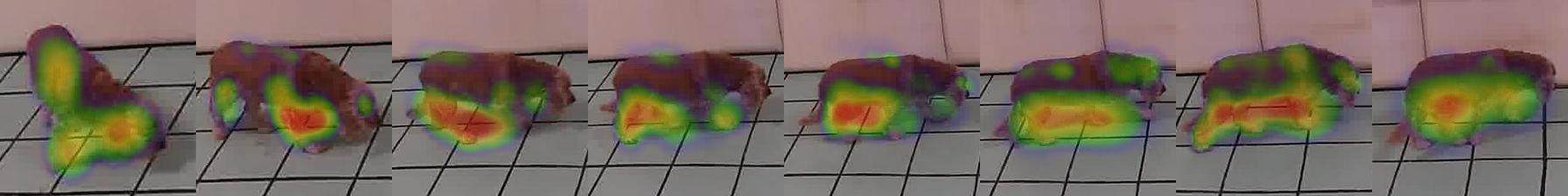}\\
 \includegraphics[width=0.8\textwidth]{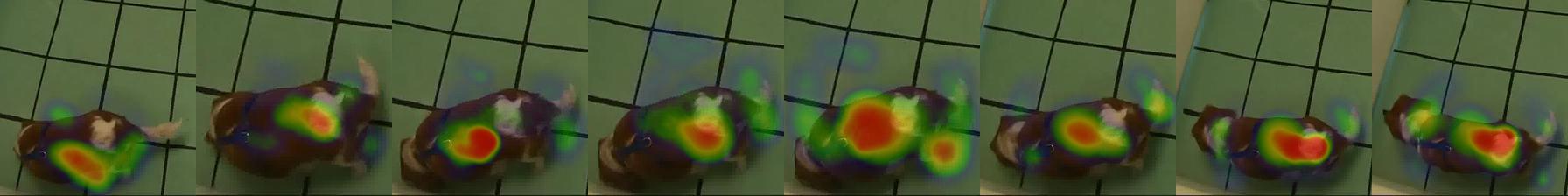}
\end{tabular}

 \caption{Saliency maps for a true positive (top) and a false negative (bottom) case belonging to the pain class.} 
 \label{fig:vis}
\end{figure}

Figure~\ref{fig:vis} shows two dogs in orthopedic pain condition. The top sequence in Figure~\ref{fig:vis} is a case that is correctly classed as pain by our ConvLSTM+LSTM model, with a confidence score of 0.95. According to the heatmap, the lower body parts and the head parts are the regions that the model most focuses on. This is reasonable, since a dog with an orthopedic pain would be unstable in their walking motion, and this would show up in their leg movements and head position. We note that the algorithm seems not to be affected too much by the image background, and could follow the temporal patterns of the dog. In the bottom sequence of Figure~\ref{fig:vis}, we see an example sequence where the dog is in pain, but the algorithm classifies it incorrectly as ``no pain", with a low confidence score. From the heatmap, we can see that algorithm is focusing on the back of the dog, and the legs are not clearly visible from this angle. Therefore, the algorithm cannot reliably detect the pain expression.

\section{Conclusions}
Recognizing pain-related behaviors in dogs automatically is a challenging problem. Clinical pain assessment methods, such as the Glasgow composite measure pain scale, use multiple behavioral indicators such as ``vocalization, attention to wound, mobility, response to touch, demeanor and posture/activity" for assessment of behaviours~\cite{reid2007development}. In this paper, we have barely scratched the surface of behavior analysis for this task, and integrated computer vision based tools into a solution that uses a small number of cues. Furthermore, we have worked only on samples with orthopedic and neuropathic pain; other categories, such as organic pain, were not included.

Our results show that both spatial-temporal features in images and keypoint based descriptions of the motion patterns are useful for the visual analysis. While data restrictions and the risk of overfitting prevented us from fine-tuning pose estimators, we show that generic tools have good performance for this task. Contrary to most work on animal pain estimation, we have not focused on the facial region. This remains a future work, but would require further clinical tools, such as validated grimace scales for labeling.
\section*{Ethical Impact Statement}
The proposed approach cannot be used as a clinical assessment tool for dog pain estimation without further tests and validation. No dogs were harmed in this research, there was no induced pain for the subjects, and ethics committee approval is obtained for the preparation of the dog video dataset. 
%\section*{Acknowledgment}
\bibliographystyle{ieeetran}
\bibliography{References}
\end{document}